\pdfoutput=1

\documentclass[11pt]{article}

\usepackage[final]{acl}

\usepackage{times}
\usepackage{latexsym}

\usepackage[T1]{fontenc}

\usepackage[utf8]{inputenc}

\usepackage{microtype}

\usepackage{inconsolata}

\usepackage{graphicx}

\usepackage{times}
\usepackage{latexsym}
\usepackage{graphicx}
\usepackage{amsmath}
\usepackage{amssymb}
\usepackage{pifont}
\usepackage{graphicx}
\usepackage{booktabs}
\usepackage{multirow}
\usepackage{xcolor}
\usepackage{color}
\usepackage{colortbl}
\usepackage[table]{xcolor}
\usepackage{tcolorbox}
\tcbuselibrary{listings, breakable}

\usepackage[T1]{fontenc}

\usepackage[utf8]{inputenc}

\usepackage{microtype}

\usepackage{inconsolata}

\title{LLM-OREF: An Open Relation Extraction Framework Based on Large Language Models}

\author{
    Hongyao Tu\textsuperscript{1,3}\thanks{Equal contribution.}\thanks{This work was partially done while Hongyao Tu was interning at the Shopee LLM Team.}~
    Liang Zhang\textsuperscript{1\footnotemark[1]}~
    Yujie Lin\textsuperscript{1}~
    Xin Lin\textsuperscript{3}~\\
    \textbf{Haibo Zhang\textsuperscript{2}~
    Long Zhang\textsuperscript{2}~
    Jinsong Su\textsuperscript{1,3}\thanks{Corresponding author.}}\\
    \textsuperscript{1}School of Informatics, Xiamen University, China, \\
    \textsuperscript{2}LLM Team, Shopee Pte. Ltd. \\
    \textsuperscript{3}National Institute for Data Science in Health and Medicine, Xiamen University \\
    \texttt{\{tuhongyao,lzhang\}@stu.xmu.edu.cn},~~~\texttt{jssu@xmu.edu.cn}  \\
}

\begin{document}
\maketitle
\begin{abstract}
The goal of open relation extraction (OpenRE) is to develop an RE model that can generalize to new relations not encountered during training. Existing studies primarily formulate OpenRE as a clustering task. They first cluster all test instances based on the similarity between the instances, and then manually assign a new relation to each cluster. However, their reliance on human annotation limits their practicality. In this paper, we propose an OpenRE framework based on large language models (LLMs), which directly predicts new relations for test instances by leveraging their strong language understanding and generation abilities, without human intervention. Specifically, our framework consists of two core components: (1) a relation discoverer (RD), designed to predict new relations for test instances based on \textit{demonstrations} formed by training instances with known relations; and (2) a relation predictor (RP), used to select the most likely relation for a test instance from $n$ candidate relations, guided by \textit{demonstrations} composed of their instances. To enhance the ability of our framework to predict new relations, we design a self-correcting inference strategy composed of three stages: relation discovery, relation denoising, and relation prediction. In the first stage, we use RD to preliminarily predict new relations for all test instances. Next, we apply RP to select some high-reliability test instances for each new relation from the prediction results of RD through a cross-validation method. During the third stage, we employ RP to re-predict the relations of all test instances based on the demonstrations constructed from these reliable test instances. Extensive experiments on three OpenRE datasets demonstrate the effectiveness of our framework. We release our code at \url{https://github.com/XMUDeepLIT/LLM-OREF.git}.
\end{abstract}

\section{Introduction}

Relation Extraction (RE), as a crucial task in information extraction, aims to extract relations between entity pairs from unstructured text. The extracted relations play a vital role in many downstream applications, such as search engine \citep{li2006relation}, question answering \citep{yu2017improved}, and knowledge base population \citep{ji2011knowledge}. Conventional RE studies  mainly focus on building models that can only handle predefined relations, inherently limiting their utility in real-world scenarios where new relations continually emerge. To address this limitation, researchers have turned to Open Relation Extraction (OpenRE), which is not confined to a predefined set of relations and can dynamically discover new ones, making it more practical for real-world applications.

In this regard, dominant methods formulate OpenRE as a clustering task \citep{yao2011structured,marcheggiani2016discrete,elsahar2017unsupervised}, which aggregates semantically related relation instances into the same cluster, with each cluster representing a potential new relation. Along this line, subsequent studies directly utilize pretrained language models (e.g., BERT \citep{devlin2019bert}) to encode an instance for obtaining its relational representation and then perform clustering on these representations \citep{hu2020selfore}. Since pretrained language models have not been fine-tuned on RE datasets, the performance of such methods remains suboptimal. To deal with this issue, several methods leverage the available labeled datasets for RE (which only contain known relations) to fine-tune pretrained language models \citep{zhao2021relation, wang2022matchprompt}. However, the above methods cannot align clusters with specific relation types, restricting their applicability in downstream tasks. While  \citet{zhao2023actively} mitigates this issue by actively selecting representative instances for human annotation during clustering, their approach remains impractical for real-world deployment due to its reliance on human intervention.

Recently, Large Language Models (LLMs) have demonstrated strong text understanding capabilities across various downstream tasks and can effectively capture complex relation patterns in text sequences \citep{wan2023gpt,wadhwa2023revisiting}. More importantly, unlike traditional classification and clustering models, the generative nature of LLMs allows them to predict new relations in the form of natural language directly. Therefore, we believe that exploring the potential of LLMs in OpenRE is a promising research direction.

In this work, we conduct a preliminary study to investigate the capabilities of LLMs in OpenRE. We first observe that  LLMs perform poorly at predicting new relations in a zero-shot manner. In addition, providing LLMs with a few-shot demonstration that includes instances that do not belong to new relations can improve their performance, but the improvement is limited. Interestingly, we find that their performance significantly improves when instances of the new relations are provided in few-shot demonstrations. These findings suggest that while LLMs excel in comprehending relations through demonstrations, they still struggle to discover new relations.

Based on these insights, we propose an LLM-based Open Relation Extraction Framework (LLM-OREF), which consists of two key components: the Relation Discoverer (RD) and the Relation Predictor (RP). The former aims to preliminarily predict new relations for test instances by capturing relation patterns from demonstrations composed of instances with known relations. The latter is designed to deeply understand the relations of instances in demonstration, and then accurately determine the most probable relation for the test instance from these relations. The primary distinction between RD and RP lies in their input: whether the demonstration contains the relation of the test instance. To reduce storage and training costs, both RD and RP are built on the same LLM using the LoRA \citep{hu2022lora} fine-tuning strategy. During training, since we can only access instances of known relations, we construct the inputs of RD and RP using these instances for corresponding training. As RP’s demonstration includes the target relation of the test instance, which greatly reduces the difficulty of relation prediction, its performance is significantly better than that of RD. Therefore, to enhance the training of RD, we introduce an extra distillation loss ($\mathcal{L}_{\text{KD}}$) in its training objective, designed to leverage the output distribution of RP to guide RD's training.

To more effectively coordinate RD and RP to discover new relations in real-world scenarios and accurately predict relations of test instances, we propose a self-correcting inference strategy. In particular, the strategy involves three stages: relation discovery, relation denoising, and relation prediction. In the first stage, we employ RD to initially predict new relations for each instance in the test set, based on demonstrations consisting of training instances with known relations. In the second stage, considering that RD's predictions may contain noise, we use RP to cross-validate the accuracy of the predicted relation for each test instance, yielding a set of reliable instances for each new relation. In the third stage, we apply RP to more accurately predict new relations for each test instance using demonstrations constructed from these reliable instances.

To summarize, the main contributions of this work are as follows: (1) We propose LLM-OREF, a novel OpenRE framework based on LLMs that includes two key components, the RD and RP, to enable the discovery of new relations and their accurate prediction. (2) We propose a self-correcting inference strategy that progressively refines new relation prediction through a three-stage pipeline of relation discovery, relation denoising, and relation prediction. (3) Extensive experiments conducted on three OpenRE datasets demonstrate the effectiveness of our framework.

\begin{figure}
\setlength{\abovecaptionskip}{4pt}
\setlength{\belowcaptionskip}{-8pt}
    \centering
    \includegraphics[width=0.48\textwidth]{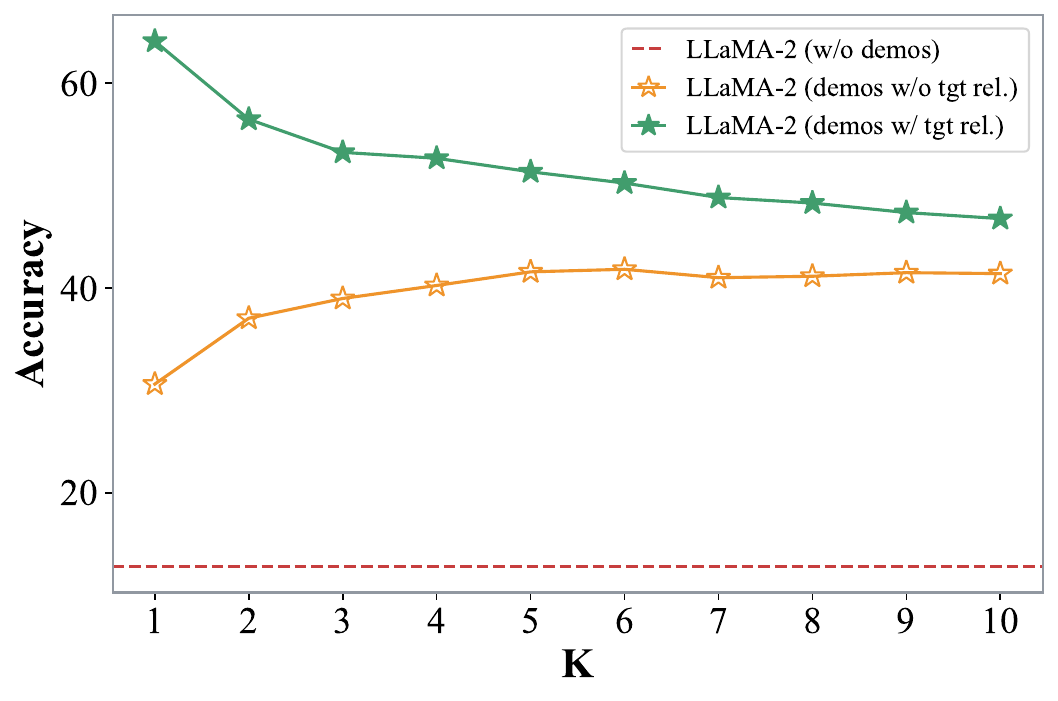}
    \caption{Accuracy of \textbf{LLaMA-2-7B} on FewRel. ``\textbf{\textit{w/o demons}}'' means no demonstrations are given; ``\textbf{\textit{demos w/o tgt rel.}}'' and ``\textbf{\textit{demos w/ tgt rel.}}'' refer to demonstrations that do not contain and that contain instances sharing the target relation of the test instance, respectively. On the x-axis, ``\textbf{\textit{K}}'' denotes the number of instances in demonstrations. In the ``demos w/ tgt rel.'' setting, as $K$ increases, it becomes harder for the model to identify the target relation, leading to a gradual drop in accuracy.}
    \label{fig:preliminary}
\end{figure}

\section{Preliminary Study}
In this section, we conduct a preliminary study to explore the ability of LLMs in discovering new relations. To this end, we evaluate the performance of an open-source LLM, LLaMA-2-7B \citep{touvron2023llama}, on a commonly used OpenRE dataset FewRel \citep{han2018fewrel}.

Specifically, we first simply evaluate the accuracy of the LLM on the test set of FewRel under a zero-shot setting. The red dashed line in Figure \ref{fig:preliminary} indicates that the LLM exhibits notably poor performance. This is mainly attributed to the fact that the model lacks task-specific guidance, having not been exposed to any examples that illustrate the structure and requirements of RE. To enhance the LLM’s understanding of the RE task, we provide it with demonstrations, including instances of known relations in the training set of FewRel, to predict new relations of test instances under a few-shot setting. While this setting improves the LLM's ability to identify new relations, its performance is still inadequate for real-world applications (see yellow-\ding{73} line in Figure \ref{fig:preliminary}). These results intuitively reveal that the ability of LLMs to discover new relations is still limited. This motivates us to further explore more effective methods and strategies to enhance the performance of LLMs on OpenRE.

Furthermore, we explore the performance of LLMs when demonstrations include the target (new) relations of test instances, as done in standard in-context learning (ICL) \citep{brown2020language}. From the green-$\bigstar$ line in Figure \ref{fig:preliminary}, we observe a substantial improvement in the LLM’s performance. These findings indicate that LLMs can effectively grasp the semantics of a relation through its instances in the demonstration, enabling them to accurately identify the most likely relation for a test instance from those presented in the demonstration. This has been noted in previous ICL-based RE studies~\citep{wan2023gpt, rajpoot2023gpt} and inspired the design of our framework.

\section{Problem Definition}
OpenRE endeavors to accurately predict target (new) relations for test instances in real-world scenarios. Hence, given an unlabeled dataset $ D$$=$$\{x_i\}_{i=1}^N $ (i.e., test set) with $N$ test instances, the OpenRE model is required to predict a new relation $y_i$ for each test instance $x_i$. Meanwhile, each instance $x_i{=}{<}s_i,h_i,t_i{>}$ consists of a sentence $s_i$, a head entity $h_i$, and a tail entity $t_i$. Following recent works \citep{zhao2021relation, wang2022matchprompt}, we use a training set $D'$$=$$\{x'_j\}_{j=1}^M$ containing $M$ instances of known relations to adapt LLMs to the RE task. Here, each training instance $x'_j{=}{<}s'_j, h'_j,t'_j{>}$ is annotated with its associated relation label $y'_j$. Notably, in OpenRE, the relations in the test set do not overlap with those in the training set.

\begin{figure}
\setlength{\abovecaptionskip}{4pt}
\setlength{\belowcaptionskip}{-8pt}
    \centering
    \includegraphics[width=0.48\textwidth]{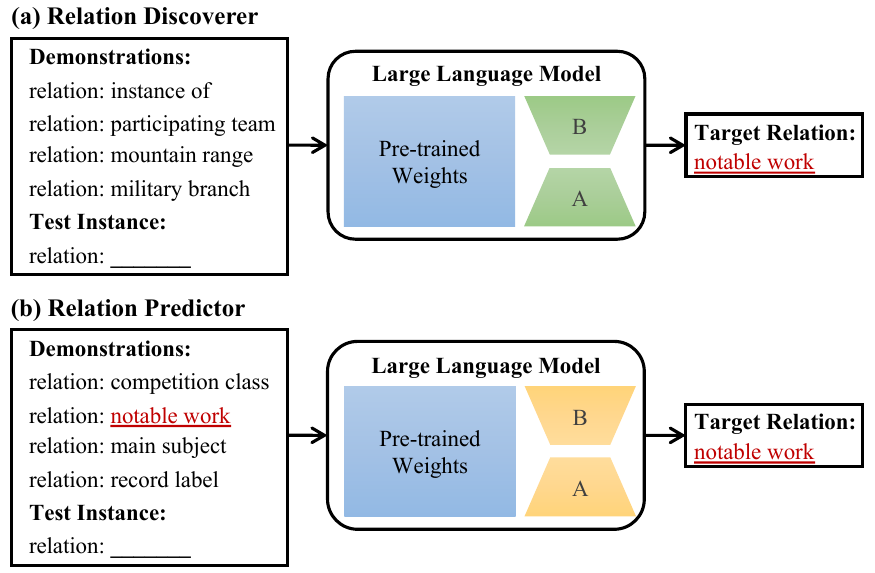}
    \caption{Illustration of two key components in our framework. The demonstrations in RD consist of known relation instances, while the demonstrations in RP are composed of new relation instances. Additionally, the demonstrations in RD do not include the target relation of the test instance, whereas RP does. Both RD and RP are built on the same LLM using the LoRA fine-tuning strategy.}
    \label{fig:components}
\end{figure}

\section{Our Framework}
In this section, we provide a detailed description of our LLM-based OpenRE framework, LLM-OREF. In the following, we first elaborate on two key components of LLM-OREF in \textsection\ref{subsec:framework}, and then detail the training and inference strategies of our framework in \textsection\ref{subsec:training} and \textsection\ref{subsec:infer}, respectively.

\subsection{Overall framework}
\label{subsec:framework}
As illustrated in Figure~\ref{fig:components}, LLM-OREF consists of two key components: the Relation Discoverer and the Relation Predictor.

\paragraph{Relation Discoverer~(RD).}
The RD endeavors to predict new relations for test instances based on demonstrations consisting of instances with known relations. 

As illustrated in Figure~\ref{fig:components}~(a), for each test instance $x_i$, we first randomly sample $n$ known relations from the training set, each associated with a training instance, and concatenate them to form demonstrations: $D_{\text{RD}}$$=$$[x'_1, y'_1, ..., x'_{\text{n}}, y'_{\text{n}}]$. Subsequently, we construct the input $I_{\text{RD}}$ for RD by concatenating the instruction prompt $P_{\text{RD}}$, the demonstrations $D_{\text{RD}}$, and the test instance $x_i$, forming $ I_{\text{RD}} $$=$$ [P_{\text{RD}}; D_{\text{RD}}; x_i]$. Here, $P_{\text{RD}}$ serves to guide the RD in comprehensively understanding the RE task through the demonstrations $D_{\text{RD}}$, while also instructing it to predict a new relation (not contained in $D_{\text{RD}}$) for the test instance $x_i$. Finally, the RD takes $I_{\text{RD}}$ as input to autoregressively generate a new relation $\hat{y}_i$ for $x_i$.

\paragraph{Relation Predictor~(RP).}
As the discovery of new relations is inherently challenging, the RD may produce noisy predictions. To mitigate this, the RP is employed to denoise and refine these predictions for the test instances. 

Specifically, as shown in Figure~\ref{fig:components}~(b), for each $x_i$ with its predicted relation $\hat{y}_i$, we first randomly select $n$$-$$1$ new relations from the prediction results of RD on the test set, each accompanied by a test instance, and additionally sample a test instance belonging to $\hat{y}_i$. Next, these instances are concatenated to form demonstrations $D_{\text{RP}}$$=$$[x_1,\hat{y}_1, ..., x_n,\hat{y}_{n}]$. Then, we create an input $I_{\text{RP}}$$=$$[P_{\text{RP}}; D_{\text{RP}}; x_i]$ for RP by concatenating a specific instruction prompt $P_{\text{RP}}$, the demonstration $D_{\text{RP}}$, and the test instance $x_i$, where $P_{\text{RP}}$ instructs RP to identify the most likely relation for $x_i$ from those contained in $D_{\text{RP}}$. Finally, we input $I_{\text{RP}}$ into RP to obtain a new predicted relation $\tilde{y}_i$ for $x_i$, which can be used to verify or refine the initial prediction $\hat{y}_i$ provided by RD.

\begin{figure}
\setlength{\abovecaptionskip}{4pt}
\setlength{\belowcaptionskip}{-8pt}
    \centering
    \includegraphics[width=0.48\textwidth]{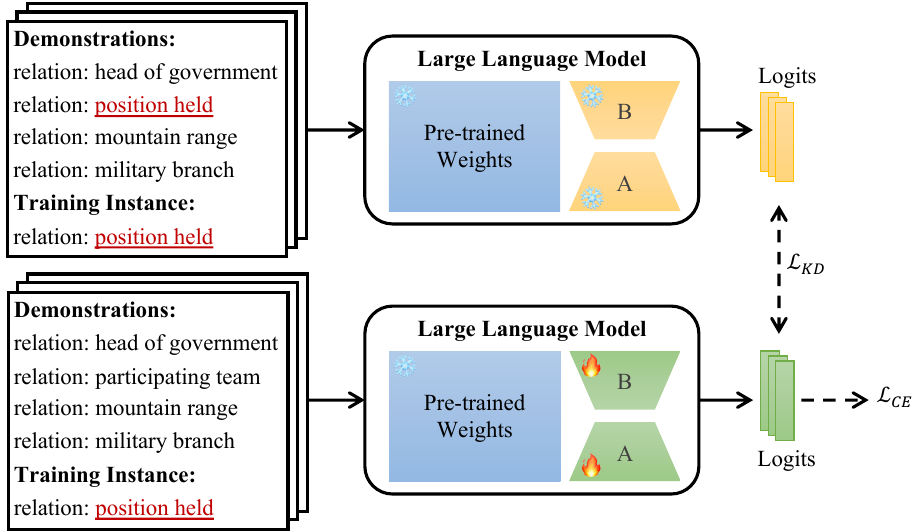}
    \caption{Illustration of the training strategy for RD. Since RP’s demonstration includes the target relation, relation prediction becomes easier, resulting in better performance than RD. To improve RD, we introduce a distillation loss ($\mathcal{L}_{\text{KD}}$) that leverages RP’s output distribution to guide RD’s training.}
    
    \label{fig:training}
\end{figure}

\begin{figure*}
\setlength{\abovecaptionskip}{4pt}
\setlength{\belowcaptionskip}{-8pt}
    \centering 
    \includegraphics[width=1\textwidth]{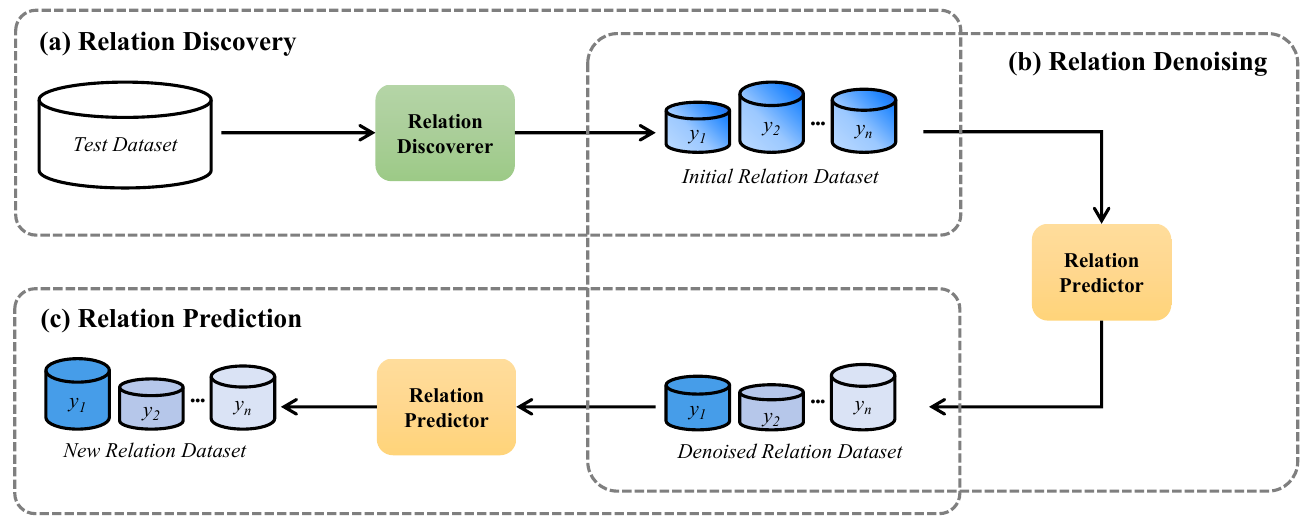}
    \caption{The illustration of our self-correcting inference strategy. It consists of three stages: (a) relation discovery, where the RD discovers potential new relations of test instances; (b) relation denoising, where the RP identifies high-reliability instances for each new relation from the prediction results of RD; and (c) relation prediction, using high-reliability instances of new relations to construct demonstrations, so that the RP can better predict relations of test instances.}
    \label{fig:inference}
\end{figure*}

\subsection{Model Training} 
\label{subsec:training}
To effectively train both RD and RP in our framework, we adopt a two-stage training strategy. For storage efficiency, we adopt LoRA to fine-tune a shared LLM for both RP and RD. Notably, during OpenRE training, the model is trained on the training set $D'$, which only contains instances of known relations. Thus, the inputs for training RP and RD are solely constructed from relations and instances sampled within $D'$.

\paragraph{The first stage.} 
Here, we focus on effectively training RP using the training set $D'$. Specifically, for each training instance $x'_j$ with its corresponding relation $y'_j$, we first randomly sample corresponding demonstrations $D_{\text{RP}}$ from $D'$. According to the objective of RP, the demonstrations $D_{\text{RP}}$ are required to include instances belonging to the relation $y'_j$. Then, we construct the input $I_{\text{RP}}$$=$$[P_{\text{RP}};D_{\text{RP}};x'_j]$ and feed it into RP. Finally, we train RP to autoregressively generate the relation $y'_j$ for the training instance $x'_j$ using the cross-entropy loss $\mathcal{L}_{\text{CE}}$:
\begin{equation}
\begin{gathered}
\mathcal{L}_{\text{CE}} = -\frac{1}{|y_j'|} \sum_{t=1}^{|y_j'|} \log P_{\boldsymbol{\theta}}( y_{j}'^{(t)} \mid  y_{j}'^{(<t)}, \mathbf{I}_{\textbf{RP}} ),
\end{gathered}
\label{loss:rp_ce}
\end{equation}
where $\boldsymbol{\theta}$ denotes the learnable LoRA parameters for RP, and $t$ is the index of a token in $y'_j$.

\paragraph{The second stage.} 
At this stage, we aim to enhance RD’s ability to discover new relations. For each training instance $x'_j$ labeled with relation $y'_j$, we construct an RD's input $I_{\text{RD}} = [P_{\text{RD}}; D_{\text{RD}}; x'_j]$, ensuring that the demonstration $D_{\text{RD}}$, sampled from $D'$, does not include any instances of $y'_j$. Next, we input $I_{\text{RD}}$ into RD and compute the autoregressive loss $\mathcal{L}'_{\text{CE}}$ with respect to the target relation $y'_j$: 
\begin{equation}
\begin{gathered}
\mathcal{L'}_{\text{CE}} = -\frac{1}{|y'_j|} \sum_{t=1}^{|y'_j|} \log P_{\boldsymbol{\phi}}( y_j'^{(t)} \mid y_{j}'^{(<t)}, \mathbf{I}_\textbf{{RD}} ),
\end{gathered}
\label{loss:rd_ce}
\end{equation}
where $\boldsymbol{\phi}$ is the learnable LoRA parameters of RD.

Since $D_{\text{RP}}$ contains instances of the target relation $y'_j$, whereas $D_{\text{RD}}$ does not, RP can more easily predict the relation $x'_j$ than RD.
Therefore, we also employ the RP trained in the first stage as a teacher to guide the training of RD. As depicted in Figure~\ref{fig:training}, we compute the KL divergence $\mathcal{L}_{\text{KD}}$ between the predictive distributions of RP and RD for the same training instance $x'_j$: 

\begin{equation}
\begin{gathered}
\mathcal{L}_{\text{KD}}= \frac{1}{|y'_j|} \sum_{t=1}^{|y'_j|}\text{\textbf{KL}}\left(P_{\boldsymbol{\theta}} ( y_j'^{(t)} \mid y_{j}'^{(<t)}, \mathbf{I}_\textbf{{RP}} ) \right. \\ \left. || P_{\boldsymbol{\phi}} ( y_j'^{(t)} \mid y_{j}'^{(<t)}, \mathbf{I}_\textbf{{RD}} ) \right).
\end{gathered}
\label{loss:rd_kl}
\end{equation}

Finally, the training objective for RD is given by $\mathcal{L}_\text{{RD}}=\mathcal{L}'_{\text{CE}} + \alpha \mathcal{L}_{\text{KD}}$, where $\alpha$ is a hyperparameter used to balance the impact of $\mathcal{L}_{\text{KD}}$ on RD training.

\subsection{Self-Correcting Inference Strategy} \label{subsec:infer}

To coordinate RD and RP to effectively discover new relations and accurately predict the relation for each test instance, we propose a self-correcting inference strategy. As depicted in Figure~\ref{fig:inference}, this strategy consists of three stages: relation discovery, relation denoising, and relation prediction.

\paragraph{Relation Discovery.} 
During this phase, RD is used to perform initial predictions of new relations for all instances in the test set $D$. 

Specifically, for each test instance $x_i$, we first construct RD's input $I_{\text{RD}}$ by randomly sampling instances with known relations from the training set $D'$ to form the corresponding demonstration $D_{\text{RD}}$. Then, we feed the input $I_{\text{RD}}$ into RD to obtain the predicted relation $\hat{y}_i$ for $x_i$. Furthermore, to improve the recall of RD in discovering new relations, we make multiple predictions for each test instance $x_i$ using different demonstrations to obtain multiple prediction relations $[\hat{y}^{1}_i, ..., \hat{y}^{K}_i]$ for each test instance $x_i$, where $K$ is the number of predictions.
 
\paragraph{Relation Denoising.} 
Here, we focus on using RP to pick out some high-reliability samples for each new relation from the prediction results of RD. 

Given each test instance $x_i$ and its predicted relation $\hat{y}^{k}_i$, we first sample multiple demonstrations $[D_{\text{RP}}^1, ..., D_{\text{RP}}^d]$, each comprising $\hat{y}^{k}_i$ and other new relations. Notably, we ensure that these demonstrations cover all new relations discovered in the previous stage. Subsequently, we utilize these sampled demonstrations to build the corresponding input $[I_{\text{RP}}^1, ..., I_{\text{RP}}^d]$ for the test instance $x_i$.
Next, we feed these inputs into RP to generate new predictions for $x_i$. This allows us to assess the reliability of $\hat{y}^{k}_i$ by comparing it to other new relations. If RP consistently outputs $\hat{y}^{k}_i$ across multiple predictions, we consider it a reliable relation for the test instance $x_i$. 

After denoising all test instances, the remaining ones undergo further rounds of denoising. In total, we perform $T$ rounds to obtain high-reliability samples.

\paragraph{Relation Prediction.}
Following the prior phase, we gathered the reliable test instances for each new relation.
In this stage, we utilize these reliable test instances to construct demonstrations $D_{\text{RP}}$ that enable RP to precisely predict new relations of other test instances.

Specifically, for a test instance $x_i$, we first sample $n$ new relations along with their reliable test instances to construct a demonstration $D_{\text{RP}}$. We then apply RP to select the most probable relation $\hat{y}_i$ for $x_i$ from these $n$ candidates. After obtaining $\hat{y}_i$, we sample other $n$$-$$1$ new relations and include $\hat{y}_i$ to form the candidate set for the next prediction. We repeat this process until all new relations have been traversed, ultimately obtaining the most reliable relation for $x_i$.

\begin{table*}[t]
\setlength{\abovecaptionskip}{4pt}
\setlength{\belowcaptionskip}{-8pt}
    \centering
    \resizebox{\linewidth}{!}{
    \begin{tabular}{ll ccc ccc c ccc}
    \toprule
    \multirow{2}{*}{\textbf{Dataset}} & \multirow{2}{*}{\textbf{Method}} & \multicolumn{3}{c}{$B^3$} & \multicolumn{3}{c}{V-measure} & \multirow{2}{*}{ARI} & \multicolumn{3}{c}{Classification}\\
    \cmidrule(lr){3-5}\cmidrule(lr){6-8}\cmidrule(lr){10-12}
    & &Prec. & Rec. & $F_1$ & Hom. & Comp. & $F_1$ & & Prec. & Rec. & $F_1$ \\
    \midrule
    \multirow{9}{*}{\textbf{FewRel}}
    &RW-HAC \citep{elsahar2017unsupervised} & 0.175 & 0.367 & 0.237 & 0.357 & 0.463 & 0.403 & 0.108 & 0.251 & 0.264 & 0.216\\
    &SelfORE \citep{hu2020selfore} &0.527  & 0.552 & 0.539 & 0.728 & 0.736 & 0.732 & 0.517 & 0.604 & 0.632 & 0.600\\
    &RSN \citep{wu2019open} & 0.174 & 0.640 & 0.274 & 0.389 & 0.659 & 0.489 & 0.173 & 0.112 & 0.239 & 0.134 \\
    &RoCORE \citep{zhao2021relation} & 0.806 & 0.843 & \underline{0.824} & 0.883  & 0.896 & 0.889 & \underline{0.807} & 0.827 & 0.868 & 0.837\\
    &ASCORE \citep{zhao2023actively}  & 0.799 & 0.841 & 0.820 & 0.888 & 0.901 & \underline{0.894} & 0.801 & 0.832 & 0.862 & \underline{0.838}\\
    &LLaMA-2-7B & 0.528 & 0.327 & 0.404 & 0.694 & 0.527 & 0.599 & 0.373 & 0.608 & 0.402  & 0.430 \\
    &Qwen2.5-14B & 0.546 & 0.555 & 0.550 & 0.725 & 0.712 & 0.719 & 0.485 & 0.710 & 0.596 & 0.586 \\
    \rowcolor{gray!20}\cellcolor{white}&\textbf{Ours (LLaMA-2-7B)}
    & 0.647 & 0.700 & 0.672 & 0.790 & 0.809 & 0.800 & 0.637 & 0.750 & 0.737 & 0.718 \\
    \rowcolor{gray!20}\cellcolor{white}&\textbf{Ours (Qwen2.5-14B)}
    &0.817 & 0.850 & \textbf{0.833} & 0.893 & 0.905 & \textbf{0.899} & \textbf{0.810} & 0.887 & 0.883 & \textbf{0.879} 
    \\ \midrule
    \multirow{9}{*}{\textbf{TACRED}} 
    &RW-HAC \citep{elsahar2017unsupervised} &0.317 & 0.668 & 0.430 & 0.443 & 0.668 & 0.532 & 0.291 & 0.244 & 0.246 & 0.171  \\
    &SelfORE \citep{hu2020selfore}  &0.517 & 0.441 & 0.476 & 0.631 & 0.600 & 0.615 & 0.434 &  0.343 & 0.396 & 0.360 \\
    &RSN \citep{wu2019open} &0.312 & 0.807 & 0.451 & 0.445 & 0.768 & 0.563 & 0.354 & 0.149 & 0.118 & 0.225\\
    &RoCORE \citep{zhao2021relation} &0.696 & 0.685 & 0.690 & 0.786 & 0.786 & 0.787 & 0.640 & 0.547 & 0.594 & 0.563 \\
    &ASCORE \citep{zhao2023actively} & 0.742 & 0.821 & \underline{0.780} & 0.807 & 0.856 & \underline{0.831} & 0.781 & 0.698 & 0.715 & \underline{0.699} \\
    &LLaMA-2-7B & 0.441 & 0.305 & 0.361 & 0.474 & 0.346 & 0.400 & 0.159 & 0.377 & 0.450 & 0.325 \\
    &Qwen2.5-14B & 0.683 & 0.610 & 0.644 & 0.713 & 0.656 & 0.684 & 0.619 & 0.709 & 0.648 & 0.592 \\
    \rowcolor{gray!20}\cellcolor{white}&\textbf{Ours (LLaMA-2-7B)}
     & 0.739 & 0.700 & 0.719 & 0.769 & 0.731 & 0.749 & \underline{0.798} & 0.665 & 0.742 & 0.633 \\
     \rowcolor{gray!20}\cellcolor{white}&\textbf{Ours (Qwen2.5-14B)}
     & 0.803 & 0.775 & \textbf{0.789} & 0.817 & 0.858 & \textbf{0.837} & \textbf{0.893} & 0.713 & 0.784 & \textbf{0.704} \\ \midrule
    \multirow{9}{*}{\textbf{FewRel-LT}}
    &RW-HAC \citep{elsahar2017unsupervised} & 0.255 & 0.322 & 0.285 & 0.379 & 0.421 & 0.399 & 0.145 & 0.190 & 0.176 & 0.160\\
    &SelfORE \citep{hu2020selfore} & 0.563 & 0.456 & 0.504 & 0.717 & 0.661 & 0.687 & 0.377 &  0.439 & 0.526 & 0.462 \\
    &RSN \citep{wu2019open} & 0.211 & 0.500 & 0.297 & 0.350 & 0.510 & 0.415 & 0.193 & 0.098 & 0.173 & 0.117\\
    &RoCORE \citep{zhao2021relation} & 0.662 & 0.717 & 0.689 & 0.800 & 0.801 & 0.800 & 0.581 & 0.507 & 0.538 & 0.517\\
    &ASCORE \citep{zhao2023actively}  & 0.650 & 0.845 & \underline{0.735} & 0.790 & 0.885 & \underline{0.835} & \underline{0.676} & 0.530 & 0.609 & 0.550 \\
    &LLaMA-2-7B  & 0.588 & 0.291 & 0.389 & 0.725 & 0.511 & 0.600 & 0.269 & 0.549 & 0.404  & 0.412 \\
    &Qwen2.5-14B & 0.601 & 0.516 & 0.555 & 0.743 & 0.679 & 0.710 & 0.478 & 0.665 & 0.595 & 0.547 \\
    \rowcolor{gray!20}\cellcolor{white}&\textbf{Ours (LLaMA-2-7B)}
     & 0.651 & 0.655 & 0.653 & 0.778 & 0.767 & 0.773 & 0.594 & 0.713 & 0.736 & \underline{0.698} \\
     \rowcolor{gray!20}\cellcolor{white}&\textbf{Ours (Qwen2.5-14B)}
     & 0.777 & 0.775 & \textbf{0.776} & 0.862 & 0.858 & \textbf{0.860} & \textbf{0.738} & 0.856 & 0.876 & \textbf{0.855} \\
    \bottomrule
    \end{tabular}
    }
    \caption{Main results on three OpenRE datasets. The experimental results demonstrate the effectiveness of our framework under both class-balanced and class-imbalanced settings.}
    \label{tab:main_res}
\end{table*}

\section{Experiments}
\subsection{Datasets \& Evaluation Metrics}
Following \citet{zhao2023actively}, we conduct experiments on two widely used RE datasets: FewRel~\citep{han2018fewrel} and TACRED~\citep{zhang2017position}, as well as a constructed RE dataset, FewRel-LT~\citep{zhao2023actively}. For each dataset, we split the relation types into disjoint sets of known and new relations. The details of datasets are in \textbf{Appendix}~\ref{appendix:dataset}.

Following \citet{zhao2023actively}, we use $\text{B}^3$ \citep{bagga1998algorithms}, $\text{V-measure}$ \citep{Rosenberg_Hirschberg_2007}, Adjusted Rand Index (ARI) \citep{hubert1985comparing}, and $\text{Macro-F}_1$ \citep{opitz2019macro} to measure the precision and recall of results, homogeneity and completeness of results, the agreement between results and true distributions, and the classification performance.

\subsection{Baselines}
We compare our LLM-OREF (based on \textbf{LLaMA-2-7B}~\citep{touvron2023llama} and \textbf{Qwen2.5-14B}~\citep{yang2025qwen3}) with two vanilla models and five representative OpenRE baselines, including: 1) \textbf{RW-HAC} \citep{elsahar2017unsupervised}, 2) \textbf{SelfORE} \citep{hu2020selfore}, 3) \textbf{RSN} \citep{wu2019open}, 4) \textbf{RoCORE} \citep{zhao2021relation}, 5) \textbf{ASCORE} \cite{zhao2023actively}. The details of these baselines are in \textbf{Appendix}~\ref{appendix:baseline}.

\subsection{Implementation Details}
For all experiments, the number of demonstrations $n$ is set to 4, and the number of relation discovery predictions $K$ for each test instance is set to 3. The number of relation denoising iterations is set to $T$$=$$3$. The prompt templates for RD and RP are in \textbf{Appendix}~\ref{appendix:prompt}. When training, we adopt AdamW~\citep{loshchilovdecoupled} as the optimizer, along with a linear learning rate schedule. All models are trained for one epoch using LoRA with $r$$=$64 and $\alpha$$=$64. The hyper-parameters, including the learning rates for RD and RP, the distillation temperature, and the corresponding loss weight $\alpha$, are listed in Table \ref{tab:hyper_p}. During inference, we utilize vLLM~\citep{kwon2023efficient} for efficient inference acceleration. All experiments are conducted on two NVIDIA H100~(80GB).

\begin{table}[h]
\setlength{\abovecaptionskip}{4pt}
\setlength{\belowcaptionskip}{-8pt}
    \centering
    \resizebox{\linewidth}{!}{
\begin{tabular}{llcccc}
\toprule
\textbf{Dataset} & \textbf{Model} & \textbf{RP lr} & \textbf{RD lr} & \textbf{Temperature $\tau$} & \textbf{Weight $\alpha$} \\ \midrule
\multirow{3}{*}{FewRel} 
 & LLaMA-2-7B & 5e-5 & 3e-6 & 4 & 0.5 \\
 & Qwen2.5-14B & 5e-5 & 3e-6 & 4 & 0.2 \\ \midrule

\multirow{3}{*}{TACRED} 
 & LLaMA-2-7B & 1e-4 & 8e-5 & 4 & 1 \\
 & Qwen2.5-14B & 7e-5 & 6e-6 & 2 & 0.9 \\ \midrule

\multirow{3}{*}{FewRel-LT} 
 & LLaMA-2-7B & 5e-5 & 3e-6 & 4 & 0.5 \\
 & Qwen2.5-14B & 6e-5 & 3e-6 & 4 & 0.1 \\ \bottomrule
\end{tabular}
}
\caption{Hyper-parameter settings.}
\label{tab:hyper_p}
\end{table}

\begin{table*}[t]
\setlength{\abovecaptionskip}{4pt}
\setlength{\belowcaptionskip}{-8pt}
    \centering
    \resizebox{\linewidth}{!}{
    \begin{tabular}{ll ccc ccc c ccc}
    \toprule
    \multirow{2}{*}{\textbf{Dataset}} & \multirow{2}{*}{\textbf{Method}} & \multicolumn{3}{c}{$B^3$} & \multicolumn{3}{c}{V-measure} & \multirow{2}{*}{ARI} & \multicolumn{3}{c}{Classification}\\
    \cmidrule(lr){3-5}\cmidrule(lr){6-8}\cmidrule(lr){10-12}
    & &Prec. & Rec. & $F_1$ & Hom. & Comp. & $F_1$ & & Prec. & Rec. & $F_1$ \\
    \midrule
    
    \rowcolor{gray!20}
    \cellcolor{white} \multirow{7}{*}{\textbf{FewRel}}
    &\textbf{Ours} 
    & 0.647 & 0.700 & \textbf{0.672} & 0.790 & 0.809 & \textbf{0.800} & \textbf{0.637} & 0.750 & 0.737 & \textbf{0.718} \\
    & 1 \textit{w/o.} self-correcting inference strategy 
    & 0.657 & 0.571 & 0.611 & 0.801 & 0.724 & 0.761 & 0.570 & 0.753 & 0.619 & 0.624 \\
    & 2 \textit{w/o.} distillation strategy 
    & 0.625 & 0.677 & 0.650 & 0.770 & 0.789 & 0.780 & 0.613 & 0.746 & 0.726 & 0.708 \\
    & 3 \textit{w/o.} relation predictor 
    & 0.639 & 0.568 & 0.601 & 0.790 & 0.722 & 0.754 & 0.559 & 0.739 & 0.595 & 0.587 \\
    & 4 \textit{w/o.} relation discoverer
    & 0.288 & 0.111 & 0.160 & 0.404 & 0.348 & 0.374 & 0.102 & 0.638 & 0.161 & 0.235   \\
    & 5 \textit{w/o.} relation denoising stage
    & 0.582 & 0.627 & 0.604 & 0.745 & 0.762 & 0.754 & 0.570 & 0.710 & 0.691 & 0.674 \\
    & 6 \textit{w/o.} relation prediction stage
    & 0.615 & 0.675 & 0.644 & 0.765 & 0.783 & 0.774 & 0.619 & 0.726 & 0.722 & 0.696 \\ 
    \midrule

    \rowcolor{gray!20}
    \cellcolor{white} \multirow{7}{*}{\textbf{TACRED}}
    &\textbf{Ours}
    & 0.739 & 0.700 & \textbf{0.719} & 0.769 & 0.731 & \textbf{0.749} & \textbf{0.798} & 0.665 & 0.742 & \textbf{0.633} \\
    & 1 \textit{w/o.} self-correcting inference strategy
    & 0.741 & 0.681 & 0.710 & 0.770 & 0.676 & 0.720 & 0.699 & 0.563 & 0.666 & 0.541 \\
    & 2 \textit{w/o.} distillation strategy
    & 0.725 & 0.680 & 0.702 & 0.752 & 0.707 & 0.729 & 0.773 & 0.650 & 0.735 & 0.625 \\
    & 3 \textit{w/o.} relation predictor
    & 0.747 & 0.539 & 0.626 & 0.774 & 0.589 & 0.669 & 0.526 & 0.659 & 0.607 & 0.513 \\
    & 4 \textit{w/o.} relation discoverer
    & 0.384 & 0.119 & 0.181 & 0.361 & 0.246 & 0.293 & 0.071 & 0.619 & 0.247 & 0.288 \\
    & 5 \textit{w/o.} relation denoising stage
    & 0.710 & 0.658 & 0.683 & 0.747 & 0.703 & 0.724 & 0.726 & 0.645 & 0.717 & 0.605 \\
    & 6 \textit{w/o.} relation prediction stage
    & 0.697 & 0.710 & 0.703 & 0.723 & 0.718 & 0.720 & 0.780 & 0.589 & 0.634 & 0.538 \\
    \midrule

    \rowcolor{gray!20}
    \cellcolor{white} \multirow{7}{*}{\textbf{FewRel-LT}}
    &\textbf{Ours}
    & 0.651 & 0.655 & \textbf{0.653} & 0.778 & 0.767 & \textbf{0.773} & 0.594 & 0.713 & 0.736 & \textbf{0.698} \\
    & 1 \textit{w/o.} self-correcting inference strategy
    & 0.725 & 0.560 & 0.632 & 0.826 & 0.704 & 0.760 & \textbf{0.615} & 0.750 & 0.629 & 0.629 \\
    & 2 \textit{w/o.} distillation strategy 
    & 0.624 & 0.623 & 0.623 & 0.753 & 0.738 & 0.745 & 0.576 & 0.699 & 0.733 & 0.692 \\
    & 3 \textit{w/o.} relation predictor
    & 0.700 & 0.553 & 0.618 & 0.809 & 0.697 & 0.749 & 0.572 & 0.704 & 0.606 & 0.579 \\
    & 4 \textit{w/o.} relation discoverer
    & 0.341 & 0.120 & 0.177 & 0.444 & 0.351 & 0.392 & 0.108 & 0.573 & 0.169 & 0.241 \\
    & 5 \textit{w/o.} relation denoising stage
    & 0.586 & 0.579 & 0.582 & 0.733 & 0.715 & 0.724 & 0.540 & 0.648 & 0.676 & 0.634 \\
    & 6 \textit{w/o.} relation prediction stage
    & 0.591 & 0.643 & 0.616 & 0.736 & 0.742 & 0.739 & 0.581 & 0.692 & 0.726 & 0.676 \\
    \bottomrule
    \end{tabular}
    }
    \caption{Ablation results on three RE datasets.}
    \label{tab:ablation}
\end{table*}

\subsection{Main Results}
Table~\ref{tab:main_res} presents the main experimental results comparing our framework with a range of baselines across three datasets. Next, we provide a detailed analysis of the results:

The results in Table~\ref{tab:main_res} show that our framework achieves superior performance on both the class-balanced dataset FewRel and the class-imbalanced datasets TACRED and FewRel-LT. It consistently outperforms all baselines, including the state-of-the-art ASCORE, while eliminating the need for human intervention. These results not only demonstrate the effectiveness of our framework but also provide valuable insights for future research in OpenRE.

Compared with traditional clustering methods such as RW-HAC, SelfORE, RSN, and RoCORE, which cannot automatically align clusters with specific relations, our framework demonstrates markedly greater practicality. It consistently outperforms these baselines across all datasets and achieves particularly large performance gains on the challenging TACRED and FewRel-LT datasets, both of which suffer from severe class imbalance. This superior performance under imbalanced conditions further validates the real-world applicability of our framework.

From Table \ref{tab:main_res}, we observe that the vanilla LLaMA-2-7B and Qwen2.5-14B exhibit consistently poor performance across all datasets, underscoring the limitations of directly applying LLMs for OpenRE. In contrast, our framework, built upon these models, achieves significant and consistent improvements, highlighting the necessity of a tailored framework for adapting LLMs to OpenRE. Moreover, these results demonstrate that our framework can be effectively applied to different LLMs, further confirming its generality.

\subsection{Ablation Study}
We further conduct ablation studies by removing various components of our framework to assess their individual contributions. Specifically, we compare our framework with the following variants in Table~\ref{tab:ablation}.

(1) \textit{w/o. self-correcting inference strategy.}
In this variant, we remove the self-correcting inference strategy from LLM-OREF, using RD for new relation discovery, which is then taken as the final prediction. As shown in line 1 of each dataset in Table~\ref{tab:ablation}, this results in a significant performance drop on three datasets. This indicates that the self-correcting inference strategy effectively coordinates RD and RP, enabling more accurate new relation prediction.

(2) \textit{w/o. distillation strategy} and \textit{w/o. relation predictor.}
In our framework, RP is used not only as a teacher model for knowledge distillation during training but also for relation denoising and prediction during inference. To evaluate its impact, we design two ablated variants: one without RP during training, and another without RP in both training and inference. As shown in lines 2 and 3 of each dataset, removing RP during training causes a performance drop, which becomes more pronounced when RP is also removed during inference. These results demonstrate that the distillation strategy enhances RD’s ability to discover new relations, and that RP plays a critical role in the overall effectiveness of our framework.

(3) \textit{w/o. relation discoverer.}
This variant uses RP directly to predict new relations based on demonstrations of known relation instances. However, Line 4 in each dataset shows that removing RD causes a notable performance drop across all datasets. An intuitive reason is that RP struggles to identify new relations when relying solely on known relation demonstrations. These results highlight the crucial role of RD in ensuring the practicality and effectiveness of our framework.

(4) \textit{w/o. relation denoising stage.}
Here, we remove the relation denoising stage from our framework and directly use the new relation instances discovered by RD as demonstrations for RP's relation prediction. This results in a significant performance drop across all three datasets (see line 5 of each dataset). This demonstrates that the denoising stage effectively selects noisy instances of new relations, thereby enabling RP to better understand new relations and accurately predict the relation of test instances.

(5) \textit{w/o. relation prediction stage.}
To assess the necessity of the relation prediction stage, this variant lets RD generate multiple candidate relations per test instance, with RP predicting the final relation from the candidate set. As shown in line 6 of each dataset, this approach consistently reduces final prediction performance, especially on TACRED. These results confirm that the relation prediction stage is essential for accurate new relation prediction.

In \textbf{Appendix}~\ref{appendix:analisys}, we further analyze the performance of relation discovery and the effect of the distillation loss weight $\alpha$.

\section{Related Work}
Conventional RE methods \citep{song2019leveraging, wu2022label, zhang2022towards, zhang2023exploring1, yue2024towards, zhang2023exploring2, zhang2023hypernetwork, zhang2025self, zhang2024multi} cannot handle the continual emergence of new relations in real-world scenarios, which motivates the development of OpenRE. Previous approaches can be divided into two categories: Tagging-based \cite{yates2007textrunner, etzioni2008open} and Clustering-based \cite{yao2011structured,marcheggiani2016discrete,elsahar2017unsupervised, wang2023improving}. Tagging-based methods extract relations by analyzing the syntactic structure of sentences, but they often overlook semantic information, making clustering-based approaches more appealing. The clustering-based approaches aim to aggregate semantically related relation instances into the same cluster, with each cluster representing a potential new relation. \citet{wu2019open} leverages labeled data from predefined relations to train a model that can measure semantic similarity between relation instances. The learned similarity metric was then applied to cluster new relation instances. With the rise of pretrained language models (e.g., BERT \citep{devlin2019bert}), many studies \citep{hu2020selfore, zhao2021relation} have leveraged these models to encode an instance for obtaining its relational representation, as they are capable of capturing deep semantic information from text. Clustering is then performed on these representations to group semantically similar relation instances. However, the clustering-based approaches above cannot align clusters with specific relation types, restricting their applicability in downstream tasks. A recent study by \citet{zhao2023actively} alleviates this issue by incorporating active learning into OpenRE, which actively selects representative instances for human annotation during the clustering process to align clusters with a specific relation type, but the need for human intervention severely restricts its practicality in real-world scenarios. Although a recent study by \citet{wang2024phrases} utilizes an API-based LLM as a phrase extractor to generate relational phrases for new relation instances, the use of a closed-source LLM incurs substantial API costs and cannot fully explore the RE ability of LLMs through training, making it difficult to distinguish fine-grained semantic relations. In this paper, we propose a novel OpenRE framework based on an open-source LLM to automatically discover new relations in real-world scenarios.

\section{Conclusion}
In this paper, we propose an LLM-based OpenRE framework, which aims to leverage the strong language understanding and generation abilities of LLMs to directly predict new relations for test instances without human intervention. The framework comprises two main components: (1) a Relation Discoverer (RD) that predicts new relations for test instances based on demonstrations built from training instances with known relations;
and (2) a Relation Predictor (RP) that identifies the most likely relation for a test instance from $n$ candidates, guided by demonstrations formed by their instances.
To improve our framework's ability to predict new relations, we introduce a self-correcting inference strategy comprising three stages: relation discovery, relation denoising, and relation prediction. Specifically, we first use RD to preliminarily predict new relations for all test instances. Then, we apply RP to select high-reliability test instances for each new relation from the prediction results of RD. Finally, we employ RP to re-predict the relations of all test instances based on demonstrations constructed from these reliable test instances. Experimental results and in-depth analysis on three public datasets demonstrate the effectiveness of our framework. 

\section*{Limitations} 
Despite its effectiveness, LLM-OREF has several limitations. First, our framework uses a fixed number of demonstrations during inference. Future studies should consider treating the number of demonstrations as a dynamic variable to better adapt to more complex scenarios in real-world applications. Second, we assume that the training data for known relations is noise-free. However, potential label noise in known relations could negatively impact new relation discovery, which future work should aim to address.

\section*{Acknowledgements}
The project was supported by the National Natural Science Foundation of China (No. 62276219), 
Natural Science Foundation of Fujian Province of China (No.2024J011001), and the Public Technology Service Platform Project of Xiamen (No.3502Z20231043). We also thank the reviewers for their insightful comments.

\bibliography{anthology}

\begin{thebibliography}{42}
\providecommand{\natexlab}[1]{#1}

\bibitem[{Bagga and Baldwin(1998)}]{bagga1998algorithms}
Amit Bagga and Breck Baldwin. 1998.
\newblock Algorithms for scoring coreference chains.
\newblock In \emph{The first international conference on language resources and evaluation workshop on linguistics coreference}, volume~1, pages 563--566. Citeseer.

\bibitem[{Brown et~al.(2020)Brown, Mann, Ryder, Subbiah, Kaplan, Dhariwal, Neelakantan, Shyam, Sastry, Askell et~al.}]{brown2020language}
Tom Brown, Benjamin Mann, Nick Ryder, Melanie Subbiah, Jared~D Kaplan, Prafulla Dhariwal, Arvind Neelakantan, Pranav Shyam, Girish Sastry, Amanda Askell, and 1 others. 2020.
\newblock Language models are few-shot learners.
\newblock \emph{Advances in neural information processing systems}, 33:1877--1901.

\bibitem[{Devlin et~al.(2019)Devlin, Chang, Lee, and Toutanova}]{devlin2019bert}
Jacob Devlin, Ming-Wei Chang, Kenton Lee, and Kristina Toutanova. 2019.
\newblock Bert: Pre-training of deep bidirectional transformers for language understanding.
\newblock In \emph{Proceedings of the 2019 conference of the North American chapter of the association for computational linguistics: human language technologies, volume 1 (long and short papers)}, pages 4171--4186.

\bibitem[{Elsahar et~al.(2017)Elsahar, Demidova, Gottschalk, Gravier, and Laforest}]{elsahar2017unsupervised}
Hady Elsahar, Elena Demidova, Simon Gottschalk, Christophe Gravier, and Frederique Laforest. 2017.
\newblock Unsupervised open relation extraction.
\newblock In \emph{The Semantic Web: ESWC 2017 Satellite Events: ESWC 2017 Satellite Events, Portoro{\v{z}}, Slovenia, May 28--June 1, 2017, Revised Selected Papers 14}, pages 12--16. Springer.

\bibitem[{Etzioni et~al.(2008)Etzioni, Banko, Soderland, and Weld}]{etzioni2008open}
Oren Etzioni, Michele Banko, Stephen Soderland, and Daniel~S Weld. 2008.
\newblock Open information extraction from the web.
\newblock \emph{Communications of the ACM}, 51(12):68--74.

\bibitem[{Han et~al.(2018)Han, Zhu, Yu, Wang, Yao, Liu, and Sun}]{han2018fewrel}
Xu~Han, Hao Zhu, Pengfei Yu, Ziyun Wang, Yuan Yao, Zhiyuan Liu, and Maosong Sun. 2018.
\newblock Fewrel: A large-scale supervised few-shot relation classification dataset with state-of-the-art evaluation.
\newblock In \emph{Proceedings of the 2018 Conference on Empirical Methods in Natural Language Processing}, pages 4803--4809.

\bibitem[{Hogan et~al.(2023)Hogan, Li, and Shang}]{hogan2023open}
William Hogan, Jiacheng Li, and Jingbo Shang. 2023.
\newblock Open-world semi-supervised generalized relation discovery aligned in a real-world setting.
\newblock In \emph{Proceedings of the 2023 Conference on Empirical Methods in Natural Language Processing}, pages 14227--14242.

\bibitem[{Hu et~al.(2022)Hu, Shen, Wallis, Allen-Zhu, Li, Wang, Wang, Chen et~al.}]{hu2022lora}
Edward~J Hu, Yelong Shen, Phillip Wallis, Zeyuan Allen-Zhu, Yuanzhi Li, Shean Wang, Lu~Wang, Weizhu Chen, and 1 others. 2022.
\newblock Lora: Low-rank adaptation of large language models.
\newblock \emph{ICLR}, 1(2):3.

\bibitem[{Hu et~al.(2020)Hu, Wen, Xu, Zhang, and Yu}]{hu2020selfore}
Xuming Hu, Lijie Wen, Yusong Xu, Chenwei Zhang, and Philip~S Yu. 2020.
\newblock Selfore: Self-supervised relational feature learning for open relation extraction.
\newblock In \emph{Proceedings of the 2020 Conference on Empirical Methods in Natural Language Processing (EMNLP)}, pages 3673--3682.

\bibitem[{Hubert and Arabie(1985)}]{hubert1985comparing}
Lawrence Hubert and Phipps Arabie. 1985.
\newblock Comparing partitions.
\newblock \emph{Journal of classification}, 2:193--218.

\bibitem[{Ji and Grishman(2011)}]{ji2011knowledge}
Heng Ji and Ralph Grishman. 2011.
\newblock Knowledge base population: Successful approaches and challenges.
\newblock In \emph{Proceedings of the 49th annual meeting of the association for computational linguistics: Human language technologies}, pages 1148--1158.

\bibitem[{Kwon et~al.(2023)Kwon, Li, Zhuang, Sheng, Zheng, Yu, Gonzalez, Zhang, and Stoica}]{kwon2023efficient}
Woosuk Kwon, Zhuohan Li, Siyuan Zhuang, Ying Sheng, Lianmin Zheng, Cody~Hao Yu, Joseph Gonzalez, Hao Zhang, and Ion Stoica. 2023.
\newblock Efficient memory management for large language model serving with pagedattention.
\newblock In \emph{Proceedings of the 29th Symposium on Operating Systems Principles}, pages 611--626.

\bibitem[{Li et~al.(2006)Li, Wang, and Huang}]{li2006relation}
Yufei Li, Yuan Wang, and Xiaotao Huang. 2006.
\newblock A relation-based search engine in semantic web.
\newblock \emph{IEEE transactions on knowledge and data engineering}, 19(2):273--282.

\bibitem[{Loshchilov and Hutter()}]{loshchilovdecoupled}
Ilya Loshchilov and Frank Hutter.
\newblock Decoupled weight decay regularization.
\newblock In \emph{International Conference on Learning Representations}.

\bibitem[{Marcheggiani and Titov(2016)}]{marcheggiani2016discrete}
Diego Marcheggiani and Ivan Titov. 2016.
\newblock Discrete-state variational autoencoders for joint discovery and factorization of relations.
\newblock \emph{Transactions of the Association for Computational Linguistics}, 4:231--244.

\bibitem[{Opitz and Burst(2019)}]{opitz2019macro}
Juri Opitz and Sebastian Burst. 2019.
\newblock Macro f1 and macro f1.
\newblock \emph{arXiv preprint arXiv:1911.03347}.

\bibitem[{Rajpoot and Parikh(2023)}]{rajpoot2023gpt}
Pawan Rajpoot and Ankur Parikh. 2023.
\newblock Gpt-finre: In-context learning for financial relation extraction using large language models.
\newblock In \emph{Proceedings of the Sixth Workshop on Financial Technology and Natural Language Processing}, pages 42--45.

\bibitem[{Rosenberg and Hirschberg(2007)}]{Rosenberg_Hirschberg_2007}
Andrew Rosenberg and Julia Hirschberg. 2007.
\newblock V-measure: A conditional entropy-based external cluster evaluation measure.
\newblock \emph{Empirical Methods in Natural Language Processing,Empirical Methods in Natural Language Processing}.

\bibitem[{Song et~al.(2019)Song, Zhang, Gildea, Yu, Wang, and Su}]{song2019leveraging}
Linfeng Song, Yue Zhang, Daniel Gildea, Mo~Yu, Zhiguo Wang, and Jinsong Su. 2019.
\newblock Leveraging dependency forest for neural medical relation extraction.
\newblock In \emph{Proceedings of the 2019 Conference on Empirical Methods in Natural Language Processing and the 9th International Joint Conference on Natural Language Processing (EMNLP-IJCNLP)}, pages 208--218.

\bibitem[{Touvron et~al.(2023)Touvron, Martin, Stone, Albert, Almahairi, Babaei, Bashlykov, Batra, Bhargava, Bhosale et~al.}]{touvron2023llama}
Hugo Touvron, Louis Martin, Kevin Stone, Peter Albert, Amjad Almahairi, Yasmine Babaei, Nikolay Bashlykov, Soumya Batra, Prajjwal Bhargava, Shruti Bhosale, and 1 others. 2023.
\newblock Llama 2: Open foundation and fine-tuned chat models.
\newblock \emph{arXiv preprint arXiv:2307.09288}.

\bibitem[{Wadhwa et~al.(2023)Wadhwa, Amir, and Wallace}]{wadhwa2023revisiting}
Somin Wadhwa, Silvio Amir, and Byron~C Wallace. 2023.
\newblock Revisiting relation extraction in the era of large language models.
\newblock In \emph{Proceedings of the conference. Association for Computational Linguistics. Meeting}, volume 2023, page 15566.

\bibitem[{Wan et~al.(2023)Wan, Cheng, Mao, Liu, Song, Li, and Kurohashi}]{wan2023gpt}
Zhen Wan, Fei Cheng, Zhuoyuan Mao, Qianying Liu, Haiyue Song, Jiwei Li, and Sadao Kurohashi. 2023.
\newblock Gpt-re: In-context learning for relation extraction using large language models.
\newblock In \emph{Proceedings of the 2023 Conference on Empirical Methods in Natural Language Processing}, pages 3534--3547.

\bibitem[{Wang et~al.(2024)Wang, Zhang, Lee, Zhong, Kang, and Liu}]{wang2024phrases}
Jiaxin Wang, Lingling Zhang, Wee~Sun Lee, Yujie Zhong, Liwei Kang, and Jun Liu. 2024.
\newblock When phrases meet probabilities: enabling open relation extraction with cooperating large language models.
\newblock In \emph{Proceedings of the 62nd Annual Meeting of the Association for Computational Linguistics (Volume 1: Long Papers)}, pages 13130--13147.

\bibitem[{Wang et~al.(2022)Wang, Zhang, Liu, Liang, Zhong, and Wu}]{wang2022matchprompt}
Jiaxin Wang, Lingling Zhang, Jun Liu, Xi~Liang, Yujie Zhong, and Yaqiang Wu. 2022.
\newblock Matchprompt: Prompt-based open relation extraction with semantic consistency guided clustering.
\newblock In \emph{Proceedings of the 2022 Conference on Empirical Methods in Natural Language Processing}, pages 7875--7888.

\bibitem[{Wang et~al.(2025)Wang, Li, Qiao, Zhou, and Li}]{wang2025towards}
Qing Wang, Yuepei Li, Qiao Qiao, Kang Zhou, and Qi~Li. 2025.
\newblock Towards a more generalized approach in open relation extraction.
\newblock Proc. of 63rd Annual Meeting of the Association for Computational Linguistics.

\bibitem[{Wang et~al.(2023)Wang, Zhou, Qiao, Li, and Li}]{wang2023improving}
Qing Wang, Kang Zhou, Qiao Qiao, Yuepei Li, and Qi~Li. 2023.
\newblock Improving unsupervised relation extraction by augmenting diverse sentence pairs.
\newblock In \emph{Proceedings of the 2023 Conference on Empirical Methods in Natural Language Processing}, pages 12136--12147.

\bibitem[{Wu et~al.(2022)Wu, Cao, Ge, Liu, Zhang, and Su}]{wu2022label}
Changxing Wu, Liuwen Cao, Yubin Ge, Yang Liu, Min Zhang, and Jinsong Su. 2022.
\newblock A label dependence-aware sequence generation model for multi-level implicit discourse relation recognition.
\newblock In \emph{Proceedings of the AAAI Conference on Artificial Intelligence}, volume~36, pages 11486--11494.

\bibitem[{Wu et~al.(2019)Wu, Yao, Han, Xie, Liu, Lin, Lin, and Sun}]{wu2019open}
Ruidong Wu, Yuan Yao, Xu~Han, Ruobing Xie, Zhiyuan Liu, Fen Lin, Leyu Lin, and Maosong Sun. 2019.
\newblock Open relation extraction: Relational knowledge transfer from supervised data to unsupervised data.
\newblock In \emph{Proceedings of the 2019 conference on empirical methods in natural language processing and the 9th international joint conference on natural language processing (EMNLP-IJCNLP)}, pages 219--228.

\bibitem[{Yang et~al.(2025)Yang, Li, Yang, Zhang, Hui, Zheng, Yu, Gao, Huang, Lv et~al.}]{yang2025qwen3}
An~Yang, Anfeng Li, Baosong Yang, Beichen Zhang, Binyuan Hui, Bo~Zheng, Bowen Yu, Chang Gao, Chengen Huang, Chenxu Lv, and 1 others. 2025.
\newblock Qwen3 technical report.
\newblock \emph{arXiv preprint arXiv:2505.09388}.

\bibitem[{Yao et~al.(2011)Yao, Haghighi, Riedel, and McCallum}]{yao2011structured}
Limin Yao, Aria Haghighi, Sebastian Riedel, and Andrew McCallum. 2011.
\newblock Structured relation discovery using generative models.
\newblock In \emph{proceedings of the 2011 conference on empirical methods in natural language processing}, pages 1456--1466.

\bibitem[{Yates et~al.(2007)Yates, Banko, Broadhead, Cafarella, Etzioni, and Soderland}]{yates2007textrunner}
Alexander Yates, Michele Banko, Matthew Broadhead, Michael~J Cafarella, Oren Etzioni, and Stephen Soderland. 2007.
\newblock Textrunner: open information extraction on the web.
\newblock In \emph{Proceedings of Human Language Technologies: The Annual Conference of the North American Chapter of the Association for Computational Linguistics (NAACL-HLT)}, pages 25--26.

\bibitem[{Yu et~al.(2017)Yu, Yin, Hasan, dos Santos, Xiang, and Zhou}]{yu2017improved}
Mo~Yu, Wenpeng Yin, Kazi~Saidul Hasan, Cicero dos Santos, Bing Xiang, and Bowen Zhou. 2017.
\newblock Improved neural relation detection for knowledge base question answering.
\newblock In \emph{Proceedings of the 55th Annual Meeting of the Association for Computational Linguistics (Volume 1: Long Papers)}, pages 571--581.

\bibitem[{Yue et~al.(2024)Yue, Lai, Yang, Zhang, Yao, and Su}]{yue2024towards}
Hao Yue, Shaopeng Lai, Chengyi Yang, Liang Zhang, Junfeng Yao, and Jinsong Su. 2024.
\newblock Towards better graph-based cross-document relation extraction via non-bridge entity enhancement and prediction debiasing.
\newblock In \emph{Findings of the Association for Computational Linguistics ACL 2024}, pages 680--691.

\bibitem[{Zhang et~al.(2023{\natexlab{a}})Zhang, Min, Su, Yu, Wang, and Chen}]{zhang2023exploring2}
Liang Zhang, Zijun Min, Jinsong Su, Pei Yu, Ante Wang, and Yidong Chen. 2023{\natexlab{a}}.
\newblock Exploring effective inter-encoder semantic interaction for document-level relation extraction.
\newblock In \emph{IJCAI}, pages 5278--5286.

\bibitem[{Zhang et~al.(2022)Zhang, Su, Chen, Miao, Zijun, Hu, and Shi}]{zhang2022towards}
Liang Zhang, Jinsong Su, Yidong Chen, Zhongjian Miao, Min Zijun, Qingguo Hu, and Xiaodong Shi. 2022.
\newblock Towards better document-level relation extraction via iterative inference.
\newblock In \emph{Proceedings of the 2022 Conference on Empirical Methods in Natural Language Processing}, pages 8306--8317.

\bibitem[{Zhang et~al.(2023{\natexlab{b}})Zhang, Su, Min, Miao, Hu, Fu, Shi, and Chen}]{zhang2023exploring1}
Liang Zhang, Jinsong Su, Zijun Min, Zhongjian Miao, Qingguo Hu, Biao Fu, Xiaodong Shi, and Yidong Chen. 2023{\natexlab{b}}.
\newblock Exploring self-distillation based relational reasoning training for document-level relation extraction.
\newblock In \emph{Proceedings of the AAAI conference on artificial intelligence}, volume~37, pages 13967--13975.

\bibitem[{Zhang et~al.(2024)Zhang, Yang, Fu, Lu, Shao, Liu, Meng, Zhou, Wang, and Su}]{zhang2024multi}
Liang Zhang, Zhen Yang, Biao Fu, Ziyao Lu, Liangying Shao, Shiyu Liu, Fandong Meng, Jie Zhou, Xiaoli Wang, and Jinsong Su. 2024.
\newblock Multi-level cross-modal alignment for speech relation extraction.
\newblock In \emph{Proceedings of the 2024 Conference on Empirical Methods in Natural Language Processing}, pages 11975--11986.

\bibitem[{Zhang et~al.(2025)Zhang, Zhang, Lu, Meng, Zhou, and Su}]{zhang2025self}
Liang Zhang, Yang Zhang, Ziyao Lu, Fandong Meng, Jie Zhou, and Jinsong Su. 2025.
\newblock A self-denoising model for robust few-shot relation extraction.
\newblock In \emph{Proceedings of the 63rd Annual Meeting of the Association for Computational Linguistics (Volume 1: Long Papers)}, pages 26782--26797.

\bibitem[{Zhang et~al.(2023{\natexlab{c}})Zhang, Zhou, Meng, Su, Chen, and Zhou}]{zhang2023hypernetwork}
Liang Zhang, Chulun Zhou, Fandong Meng, Jinsong Su, Yidong Chen, and Jie Zhou. 2023{\natexlab{c}}.
\newblock Hypernetwork-based decoupling to improve model generalization for few-shot relation extraction.
\newblock In \emph{Proceedings of the 2023 conference on empirical methods in natural language processing}, pages 6213--6223.

\bibitem[{Zhang et~al.(2017)Zhang, Zhong, Chen, Angeli, and Manning}]{zhang2017position}
Yuhao Zhang, Victor Zhong, Danqi Chen, Gabor Angeli, and Christopher~D Manning. 2017.
\newblock Position-aware attention and supervised data improve slot filling.
\newblock In \emph{Conference on empirical methods in natural language processing}.

\bibitem[{Zhao et~al.(2021)Zhao, Gui, Zhang, and Zhou}]{zhao2021relation}
Jun Zhao, Tao Gui, Qi~Zhang, and Yaqian Zhou. 2021.
\newblock A relation-oriented clustering method for open relation extraction.
\newblock In \emph{Proceedings of the 2021 Conference on Empirical Methods in Natural Language Processing}, pages 9707--9718.

\bibitem[{Zhao et~al.(2023)Zhao, Zhang, Zhang, Gui, Wei, Peng, and Sun}]{zhao2023actively}
Jun Zhao, Yongxin Zhang, Qi~Zhang, Tao Gui, Zhongyu Wei, Minlong Peng, and Mingming Sun. 2023.
\newblock Actively supervised clustering for open relation extraction.
\newblock In \emph{Proceedings of the 61st Annual Meeting of the Association for Computational Linguistics (Volume 1: Long Papers)}, pages 4985--4997.

\end{thebibliography}

\appendix

\section{Details of Experiment Setup}
\subsection{Datasets}
\label{appendix:dataset}
\paragraph{FewRel.} 
It consists of 80 relation types, with 700 instances per relation. The first 40 relations are categorized into the known relation set, while the remaining 40 are categorized into the new relation set.

\paragraph{TACRED.} 
It covers 41 relation types, in which the first 20 relations are categorized into the known relation set, while the remaining 21 are categorized into the new relation set.

\paragraph{FewRel-LT.} 
Since FewRel is a class-balanced dataset that fails to reflect the long-tail distribution of relations in real-world scenarios, we follow \citet{zhao2023actively} to construct the FewRel-LongTail~(FewRel-LT) dataset. It shares the same split of known and new relations as FewRel, with each known relation keeping 700 instances. However, the number of instances for each new relation is adjusted to $n=\frac{700}{0.5*id+1}$, where $id$ is from 0 to 39, representing each new relation.

\subsection{Baselines}
\label{appendix:baseline}
We compare our LLM-OREF with these OpenRE baselines: 1) \textbf{RW-HAC} \citep{elsahar2017unsupervised} proposes an unsupervised method for OpenRE by re-weighting word embeddings based on dependency paths and clustering the resulting relation representations. 2) \textbf{SelfORE} \citep{hu2020selfore} proposes a self-supervised framework that iteratively clusters contextualized entity pair representations using adaptive soft clustering, and refines them through a relation classification module trained with pseudo labels. 3) \textbf{RSN} \citep{wu2019open} learns similarity metrics of relations from labeled data of predefined relations, and then transfers the relational knowledge to identify new relations in unlabeled data. 4) \textbf{RoCORE} \citep{zhao2021relation} leverages the labeled data of predefined relations to learn a relation-oriented representation, while jointly optimizing objectives on both labeled and unlabeled data to improve new relation discovery. 5) \textbf{ASCORE} \cite{zhao2023actively} proposes an actively supervised clustering method for OpenRE, where clustering learning and human labeling are alternately performed, and an active labeling strategy is designed to select representative instances for labeling while dynamically discovering new relational clusters.

\subsection{Prompt Template}
\label{appendix:prompt}
Specifically, the prompt format used for both the Relation Discoverer and Relation Predictor is as follows:
\begin{tcolorbox}[colback=gray!10, colframe=gray!90, title=Prompt Template, breakable]
You are an expert in relationship extraction.\\
Consider the following relationships to extract the relationship between the head entity and the tail entity from the text.\\
The relationship must be in these possible relationships: \textcolor{brown}{[Relation Names]}. \\
\textbf{Demonstrations:} \\
\textbf{text:} \textcolor{brown}{[Text of Demo]} \\
\textbf{head\_entity:} \textcolor{brown}{[Head entity of Demo]} \\
\textbf{tail\_entity:} \textcolor{brown}{[Tail entity of Demo]} \\
\textbf{relationship:} \textcolor{brown}{[Relationship of Demo]} \\
\ldots

\textbf{text:} \textcolor{brown}{[Text of test instance]} \\
\textbf{head\_entity:} \textcolor{brown}{[Head entity of test instance]} \\
\textbf{tail\_entity:} \textcolor{brown}{[Tail entity of test instance]} \\
\textbf{relationship:} \\
\end{tcolorbox}

\section{Analysis}
\label{appendix:analisys}

\subsection{The Performance of Relation Discovery}

\begin{table}[h]
\setlength{\abovecaptionskip}{4pt}
\setlength{\belowcaptionskip}{-8pt}
    \centering
    \resizebox{\linewidth}{!}{
    \begin{tabular}{ll ccccc}
    \toprule
    \multirow{1}{*}{\textbf{Dataset}} & \multirow{1}{*}{\textbf{Method}} & \multirow{1}{*}{\textbf{Precision}} & \multirow{1}{*}{\textbf{Recall}} & \multirow{1}{*}{\textbf{Macro-$\text{F}_1$}} & \multirow{1}{*}{\textbf{Accuracy}} & \multirow{1}{*}{\textbf{Pass@K}} \\
    \midrule
    \multirow{4}{*}{\textbf{FewRel}}
    ~ & LLM-OREF & 0.750 & 0.737 & 0.718 & 0.737 & - \\ 
    ~ & RD(K=1) & 0.753 & 0.619 & 0.624 & 0.619 & 0.619 \\ 
    ~ & RD(K=3) & 0.501 & 0.401 & 0.400 & 0.396 & 0.788 \\ 
    ~ & RD(K=5) & 0.474 & 0.386 & 0.388 & 0.392 & 0.843 \\  
    \midrule
    \multirow{4}{*}{\textbf{TACRED}} 
    ~ & LLM-OREF & 0.665 & 0.742 & 0.633 & 0.757 & - \\ 
    ~ & RD(K=1) & 0.563 & 0.666 & 0.541 & 0.625 & 0.625 \\ 
    ~ & RD(K=3) & 0.425 & 0.395 & 0.349 & 0.431 & 0.791 \\ 
    ~ & RD(K=5) & 0.332 & 0.300 & 0.258 & 0.271 & 0.842 \\ 
    \midrule
    \multirow{4}{*}{\textbf{FewRel-LT}}
    ~ & LLM-OREF & 0.713 & 0.736 & 0.698 & 0.712 & - \\ 
    ~ & RD(K=1) & 0.750 & 0.629 & 0.629 & 0.607 & 0.607 \\ 
    ~ & RD(K=3) & 0.441 & 0.403 & 0.380 & 0.392 & 0.768 \\ 
    ~ & RD(K=5) & 0.448 & 0.406 & 0.384 & 0.423 & 0.833 \\ 
    \bottomrule
    \end{tabular}
    }
    \caption{Performance of Relation Discovery under different numbers of predictions $K$.}
    \label{tab:RD_analysis}
\end{table}

In the relation discovery stage, we make multiple predictions for each test instance to obtain multiple prediction relations for improving the recall of the Relation Discoverer (RD) in discovering new relations. As shown in Table~\ref{tab:RD_analysis}, the Pass@K metric significantly improves with increasing number of predictions $K$, indicating that the relations of test instances are increasingly recalled correctly by the RD. Such improved recall is critical for enabling more effective relation denoising in the subsequent stage. Therefore, we set the number of predictions $K=3$ across all datasets. Furthermore, our framework consistently outperforms the RD on all evaluation metrics, further demonstrating the effectiveness of our approach.

\subsection{The Effect of Distillation Loss Weight $\alpha$}

\begin{figure}[htbp]
\setlength{\abovecaptionskip}{4pt}
\setlength{\belowcaptionskip}{-8pt}
    \centering
    \includegraphics[width=0.48\textwidth]{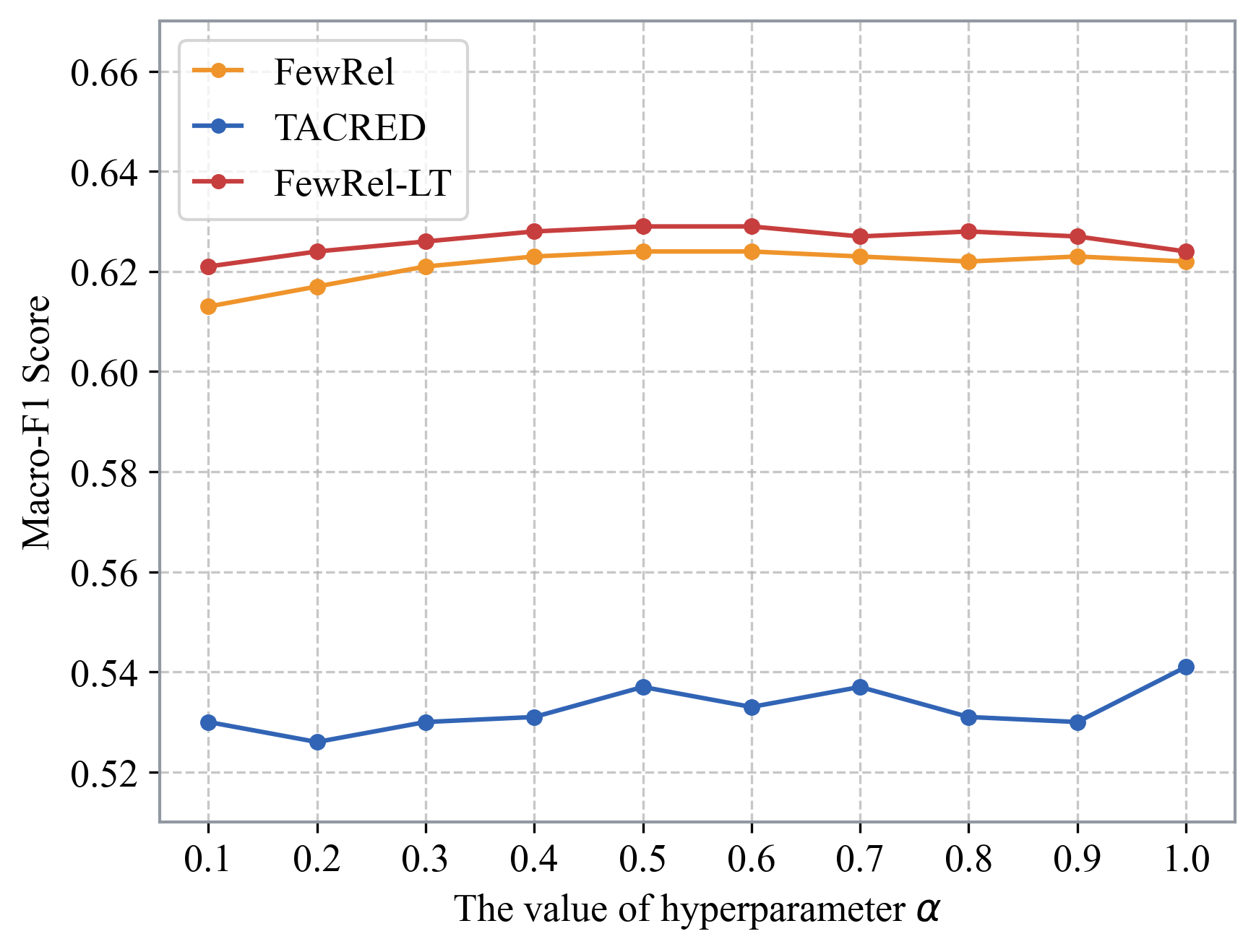}
    \caption{Performance with different weight $\alpha$ of $\mathcal{L}_{\text{KD}}$ for Relation Discoverer.}
    \label{fig:alpha_analisys}
\end{figure}

To investigate the impact of the distillation loss weight $\alpha$ on the ability of the RD to discover new relations, experiments are conducted to compare the performance of the RD by varying the value of $\alpha$. Figure \ref{fig:alpha_analisys} shows that the RD achieves the best performance on the FewRel and FewRel-LT when $\alpha=0.5$, and on the TACRED when $\alpha=1$. Meanwhile, we observe that the distillation strategy exhibits low sensitivity to the value of $\alpha$. Therefore, we set $\alpha=0.5$ for the FewRel and FewRel-LT, and $\alpha=1$ for the TACRED.

\section{Discussion}
\subsection{OpenRE Setting}

\begin{table}[h]
\setlength{\abovecaptionskip}{4pt}
\setlength{\belowcaptionskip}{-8pt}
    \centering
    \resizebox{\linewidth}{!}{
    \begin{tabular}{ll cccc}
    \toprule
    \multirow{1}{*}{\textbf{Dataset}} & \multirow{1}{*}{\textbf{Method}} & \multirow{1}{*}{\textbf{$\text{F}_1$}} & \multirow{1}{*}{$\mathbf{B^3\text{-}F_1}$}  & \multirow{1}{*}{\textbf{V-measure $\text{F}_1$}} & \multirow{1}{*}{\textbf{ARI}} \\
    \midrule
    
    \multirow{3}{*}{\textbf{FewRel}}
    ~ & KNoRD \citep{hogan2023open} & 0.774 & 0.732 & 0.730 & 0.695 \\ 
    ~ & MixORE \citep{wang2025towards} & 0.833 & 0.897 & 0.880 & 0.882 \\ 
    ~ & Ours & \textbf{0.941} & \textbf{0.898} & \textbf{0.902} & \textbf{0.883} \\ 
    \midrule

    \multirow{3}{*}{\textbf{TACRED}} 
    ~ & KNoRD \citep{hogan2023open} & 0.852 & 0.768 & 0.788 & 0.719 \\ 
    ~ & MixORE \citep{wang2025towards} & 0.883 & 0.868 & 0.860 & 0.847 \\ 
    ~ & Ours & \textbf{0.907} & \textbf{0.868} & \textbf{0.867} & \textbf{0.871} \\ 
    \midrule

    \multirow{3}{*}{\textbf{FewRel-LT}}
    ~ & KNoRD \citep{hogan2023open} & 0.867 & 0.639 & 0.731 & 0.509 \\ 
    ~ & MixORE \citep{wang2025towards} & 0.916 & 0.875 & 0.861 & 0.893 \\ 
    ~ & Ours & \textbf{0.959} & \textbf{0.890} & \textbf{0.896} & \textbf{0.898} \\ 
    \bottomrule
    \end{tabular}
    }
    \caption{Performance of Relation Discovery under different numbers of predictions $K$.}
    \label{tab:mixed_setting}
\end{table}
Recent studies \citep{hogan2023open, wang2025towards} suggest that the unlabeled dataset is typically a mixture of known and new relations. For a fair comparison, we adopt the test setting used by most existing works, where the test set contains only instances of new relations. This commonly used setting is generally more challenging than the mixed setting, as RE models typically perform better on relations they have encountered during training. Consequently, the performance of RE models under this widely used setting can be regarded as a lower bound of their performance in the mixed setting. Here, we further compare our method with the relevant baselines under the mixed setting. From Table~\ref{tab:mixed_setting}, we observed that our method still outperforms all baselines under this setting, further validating the effectiveness and robustness of our approach.

\end{document}